\documentclass[letterpaper, 10 pt, conference]{ieeeconf}  

\IEEEoverridecommandlockouts                              

\newcommand{\todo}[1]{}
\renewcommand{\todo}[1]{{\color{red} TODO: {#1}}}
\usepackage[bookmarks=true]{hyperref}

\title{\LARGE \bf
SGTM 2.0: Autonomously Untangling Long Cables\\ 
using Interactive Perception
}

\author{Kaushik Shivakumar$^{*1}$, Vainavi Viswanath$^{*1}$, Anrui Gu$^{1}$, Yahav Avigal$^{1}$, Justin Kerr$^{1}$, \\
Jeffrey Ichnowski$^{1}$, Richard Cheng$^{2}$, Thomas Kollar$^{2}$, Ken Goldberg$^{1}$
\thanks{$^{*}$ Equal Contribution}%
\thanks{$^{1}$ AUTOLAB at University of California, Berkeley}%
\thanks{$^{2}$ Toyota Research Institute (TRI)}%
}

\let\labelindent\relax
\usepackage[usenames, dvipsnames, svgnames,table,xcdraw]{xcolor}
\usepackage{graphicx}
\usepackage{algpseudocode}
\usepackage{algorithm}
\usepackage{amssymb}
\usepackage{amsmath}
\usepackage[shortlabels]{enumitem}
\usepackage{bbm}

\newcommand{\algname}{Sliding and Grasping for Tangle Manipulation 2.0 (SGTM 2.0)}

\newcommand{\algabbr}{SGTM 2.0}

\DeclareMathOperator*{\argmax}{arg\,max}

\begin{document}

\maketitle
\thispagestyle{empty}
\pagestyle{empty}

\begin{abstract}
Cables are commonplace in homes, hospitals, and industrial warehouses and are prone to tangling.
This paper extends prior work on autonomously untangling long cables by introducing novel uncertainty quantification metrics and actions that interact with the cable to reduce perception uncertainty. We present \algname{}, a system that autonomously untangles cables approximately 3 meters in length with a bilateral robot using estimates of uncertainty at each step to inform actions. By interactively reducing uncertainty, \algname{} reduces the number of state-resetting moves it must take, significantly speeding up run-time. Experiments suggest that \algabbr{} can achieve 83\% untangling success on cables with 1 or 2 overhand and figure-8 knots, and 70\% termination detection success across these configurations, outperforming SGTM 1.0 by 43\% in untangling accuracy and 200\% in full rollout speed. Supplementary material, visualizations, and videos can be found at \href{https://sites.google.com/view/sgtm2}{sites.google.com/view/sgtm2 }.


\end{abstract}


\section{Introduction}
Long cables, including electrical cords, ropes, and string, are ubiquitous in households and industrial settings \cite{mayer2008system,sanchez2018robotic,van2010superhuman}. These single-dimensional deformable objects can form knots that may restrict functionality or create hazards. Furthermore, as cable length increases, perception and manipulation of these objects become more difficult as the increased amount of free cable (which we refer to as \textit{slack}) can cause the cable to not only fall into unreachable areas of the workspace, but also form complex knots and reach irrecoverable states. Further, retrieving full state information from image observations is especially challenging when slack occludes or falsely resembles true knots.

In our prior work, we approached partial observability by using manipulation primitives which attempt to simplify state for perception modules \cite{viswanath2022autonomously}. However, this approach lacked uncertainty awareness and took actions that were often exceedingly aggressive or conservative (see \ref{subsec:prior-cable-untangling}). To address this limitation, this paper focuses on quantifying uncertainty to enable applying interactive perception \cite{bohg2017interactive}, which involves physically manipulating objects in a scene to better understand it. By considering perceptual uncertainty, we are able to perform targeted interactive perception actions that clarify the state, allowing us to perform subsequent actions with higher confidence.


\begin{figure}[ht!]
\centering
\includegraphics[width=1.0\linewidth]{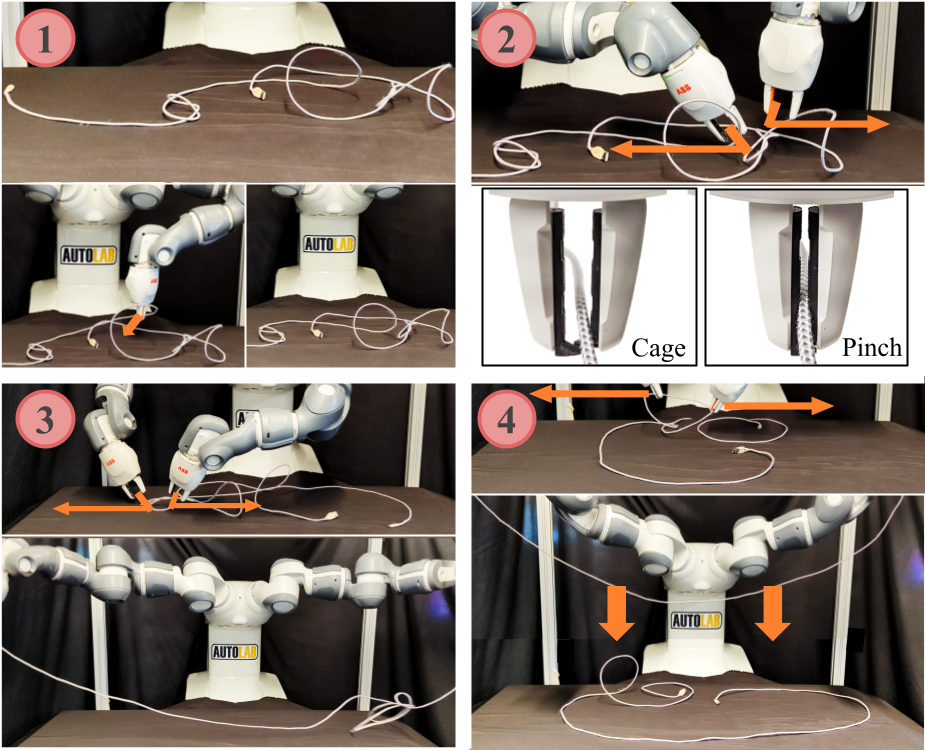}
\vspace*{-0.25in}
\caption{\textbf{Overview of \algname{}}: \algabbr{} untangles a long cable with 2 figure-8 knots. (1) The system cannot perceive a clear path to a knot and performs an exposure move, bringing the endpoint cable segment back into the observable workspace. (2) \algabbr{} confidently untangles the figure-8 knot using cage-pinch dilation. (3) The system untangles the last figure-8 knot in the scene and does an incremental Reidemeister move. (4) \algabbr{} perceives a knot-like region and uses a partial cage-pinch dilation to disambiguate it. After another incremental Reidemeister move, the system terminates confidently having verified that no knots remain.} \vspace{-0.2in}
\label{fig:splash}
\end{figure}
This paper makes the following contributions over SGTM 1.0 \cite{viswanath2022autonomously}:
\begin{enumerate}
    \item Novel perception-based metrics to estimate untangling-specific uncertainty in cable configurations, including tracing, network, and observational uncertainties as described in Section \ref{sec:IP_systems}.
    \item Novel primitives, including interactive perception actions, for cable slack management, untangling, and termination described in Section \ref{sec:manip-primitives} to reduce the probability of irrecoverable failures.
    \item \algabbr{}, an algorithm using uncertainty quantification to parameterize interactive perception actions for untangling described in Section \ref{subsec:algorithm_flow} (overview in Figure \ref{fig:splash}).
    \item Data from physical experiments suggesting 43\% improvement in untangling accuracy and 200\% improvement in speed compared to SGTM 1.0, and data suggesting that interactive perception improves accuracy by 21\% in complex cases.
\end{enumerate}

\section{Related Work}
\subsection{Deformable Object Manipulation}
Autonomous deformable object manipulation is an open problem in robotics. Deformable objects have infinite-dimensional configuration spaces, are prone to self-occlusions, and are subject to complex dynamics.
An increasingly popular approach to these problems is end-to-end deep learning with imitation learning \cite{seita2019deep, seita2020learning, nair2017combining} or reinforcement learning \cite{matas2018sim, wu2019learning, lee2020learning}. Since large-scale physical data collection is difficult, one technique is training in simulation and deploying the learned policy on physical systems~\cite{matas2018sim, wu2019learning, seita2020deep, seita2021learning, yan2020learning, sundaresan2020learning, ganapathi2021learning, hoque2020visuospatial, zhang2021robots, limr2s2r2022}. However, there the sim-to-real gap remains large due to challenges in modeling deformable objects \cite{seita2020deep}. An alternative approach is to use self-supervised learning to collect the training data directly on the physical system~\cite{avigal2022speedfolding, lee2020learning, chenfling2022}.

In multi-step algorithmic pipelines, perception-based techniques have shown to be effective for deformable object manipulation. Prior work uses dense object descriptors \cite{florence2018dense} 
for cable knot tying \cite{sundaresan2020learning} and fabric smoothing \cite{ganapathi2020learning}, 
as well as visual dynamics models for non-knotted cables \cite{yan2020learning, wang2019learning2} and fabric \cite{hoque2020visuospatial, yan2020learning, lin2022learning}. However, robust cable state estimation and dynamics estimation remain challenging for self-occluded, knotted configurations. We build on prior keypoint-based work \cite{viswanath2022autonomously, viswanath2021disentangling, grannen2020untangling} with uncertainty-aware primitives and interactive perception for the task of autonomously untangling long cables. 

\subsection{Cable Untangling}
\label{subsec:prior-cable-untangling}
Early work by Lui and Saxena \cite{lui2013tangled} uses traditional feature extraction to represent a cable's structure as a graph. Recent work \cite{grannen2020untangling, sundaresan2021untangling, viswanath2022autonomously} use learning-based keypoint detection to parameterize action primitives in an untangling pipeline. Specifically, we improve upon Sliding and Grasping for Tangle Manipulation (SGTM 1.0) \cite{viswanath2022autonomously}, an algorithm built on action primitives for autonomous long cable untangling. However, SGTM 1.0 may proceed with untangling despite low confidence in the predicted actions, leading to irrecoverable states. It also relies on shaking, a randomized reset primitive, when progress is difficult. Finally, it also requires a time-consuming physical trace to achieve the necessary confidence for termination. \algabbr{} addresses these three issues through \textit{interactive} perception, by taking untangling actions sensitive to uncertainty, making targeted moves to reduce uncertainty instead of generic recovery moves, and terminating only once sufficiently certain across multiple views that the cable is untangled.

\subsection{Active and Interactive Perception}
In 1988, Bajcsy~\cite{bajcsy1988active} defined \textit{active perception} as a search of models and control strategies for perception. Strategies vary according to the sensor and the task goal, including controlling camera parameters~\cite{bajcsy2018revisiting} and moving a tactile sensor according to haptic input~\cite{goldberg1984active}. Recently, Bohg et al.~\cite{bohg2017interactive} explore the differences between \textit{active} and \textit{interactive} perception, the latter of which specifically exploits environment interactions to simplify and enhance perception to achieve a better understanding of the scene~\cite{bohg2017interactive, novkovic2020object}.
Within robotic manipulation, several works have focused on improving understanding of the environment through scene interaction. Tsikos and Bajcsy~\cite{tsikos1988segmentation} propose interacting with random heaps of unknown objects through pick and push actions for scene segmentation. Danielczuk et al.~\cite{danielczuk2019mechanical} present the mechanical search problem, where a robot locates and retrieves an occluded target object from a cluttered bin through a series of targeted parallel jaw grasps, suction grasps, and pushes. Novkovic et al.~\cite{novkovic2020object} propose a combination of camera motions with environment interactions to find a target cube hidden in a pile of cubes. 

Interactive perception has also been applied to deformable manipulation. Willimon et al.~\cite{willimon2011classification} interact with a pile of laundry to isolate and identify individual clothing items. In our work, the robot interacts with the cable to reveal more information about the cable state.
\section{Problem Statement}
As in \cite{viswanath2022autonomously}, we consider untangling long ($\sim3$\,m) cables from RGB-D image observations. We use a bimanual robot to execute manipulation primitives until the cable reaches a fully untangled state with no knots.


\subsection{Workspace Definition and Assumptions}
The bilateral robot operates in an $(x, y, z)$ Cartesian coordinate frame with two 6-DOF robot arms. The robot is equipped with cage-pinch jaws introduced in \cite{viswanath2022autonomously} to allow for both sliding along and tightly pinching the cable (Figure \ref{fig:splash}(2)). The workspace lies in the $xy$-plane and the only inputs to the algorithm consist of RGB-D images.
The workspace contains a single incompressible electrical cable of length $l_c$ and cross-sectional radius $r_c$. Cable state $s \in \mathcal{S}$ can be described as a continuous path $c_n(u): [0, 1] \rightarrow (x, y, z)$ in the workspace, where $u$ indexes the position along the length of the cable. $c_n(0)$ and $c_n(1)$ always refer to the position of the endpoints of the cable. We initialize the cable's state before $n=0$ with the procedures specified in Section \ref{sec:expts}. One challenge in this problem is that parts of the cable may rest outside the reachable and observable workspace at any point in a rollout (defined as a single experiment aiming to remove all knots in the cable). Moreover, self-occlusions in the cable are possible due to only one overhead camera view. This partial observability motivates the need for actions that reveal more information about the cable state $s$. 

We make the following assumptions: (1) the cable can be segmented from the background via color thresholding; (2) transformations between the camera, workspace, and robot frames are known; and (3) the cable start state contains overhand or figure 8 knots of dense (6-8 cm diameter) or loose (12-14 cm diameter) configurations in series. 

\begin{figure*}[!ht]
    \centering
    \includegraphics[width=0.9\linewidth]{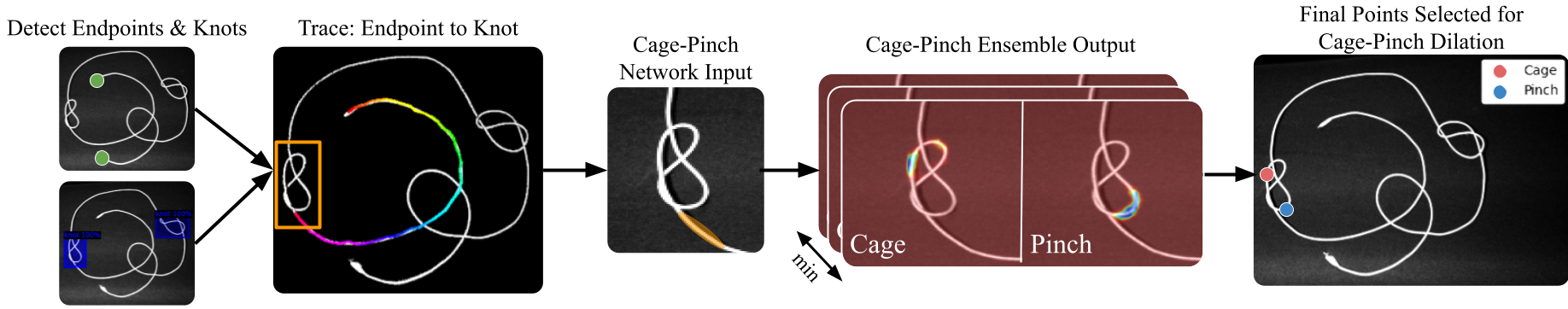}
    \caption{\textbf{Perception system:} This is the pipeline used to determine which points to cage and pinch for a cage-pinch dilation move, the crucial action for untangling a knot. First, we detect the endpoints and knots. Next, we trace from the endpoint to the first bounding box. If the trace is certain, we run the cage-pinch network ensemble on the cropped knot in the bounding box with the trace tail encoded into one of the channels. We take the pixelwise minimum across the cage-pinch network ensemble outputs, leading to 1 heatmap encoding  ``worst-case'' untangling success probabilities each for the cage and pinch point. We take the $\argmax$ of each of the two heatmaps to determine the final points to pinch and cage during the cage-pinch dilation action.}
    \vspace*{-0.2in}
    \label{fig:cagepinch-example}
\end{figure*}

\subsection{Task Objective and Metrics}
The goal of the robot is to untangle the cable and terminate at time $t < T_{\max}$, specified in Section \ref{sec:expts}. After each step of a rollout $r$, a new observation $o$ of the cable state $s$ is taken. Each primitive constitutes at least one step.

The goal of the robot over the course of each rollout is to untangle the cable and output a termination signal (\texttt{DONE}). We use $H_\texttt{DONE}$ to represent a step function, with the step occurring when the robot outputs \texttt{DONE}. We define an untangled cable as one that has no knots when its endpoints are grasped top-down and extended the maximum feasible distance, with knots defined identically to \cite{viswanath2022autonomously}. We use $k^r_t$ to denote the number of knots in the cable at time $t$ in rollout $r$ and assume the cable is initialized with $k^r_0$ knots.

We use the following metrics to measure performance, where $0 < K \leq k^r_0$ refers to the number of knots untangled and $R$ is the total set of rollouts:
\begin{enumerate}
    \item \textit{Untangling K Success Rate,} the percentage of rollouts that untangle $K$ knots: $\frac{1}{\lvert R \rvert} \sum_{r \in R} \mathbf{1}_{\{\exists t < T_{\max} \ : \ k_t^r \leq k^r_0 - K\}}$ \vspace{0.05in}
    \item \textit{Untangling Verification Rate}, the percentage of rollouts that untangle all knots and terminate successfully: $\frac{1}{\lvert R \rvert} \sum_{r \in R} \mathbf{1}_{\{\exists t < T_{\max} \ : \ k_t^r = 0 \ \land \ \texttt{DONE}_t\}}$ \vspace{0.05in}
    \item \textit{Average Untangling K Time}, the average time to untangle $K$ knots across all applicable rollouts $R_a$ where this occurs before $T_{\max}$: $\frac{1}{\lvert R_a \rvert} \sum_{r \in R_a} (\min t \ : \ k^r_t \leq k^r_0 - K )$ \vspace{-0.1in}
    \item \textit{Average Untangling Verification Time,} the average time to reach $k^r_t=0$ and declare termination across all applicable rollouts $R_a$ like above: $\frac{1}{\lvert R_a \rvert} \sum_{r \in R_a} (\min t \ : \ k^r_t = 0 \land \texttt{DONE}_t)$
\end{enumerate}
\section{Methods}

\subsection{Approach Overview}
Unlike SGTM 1.0, \algabbr{} uses interactive perception primitives designed to better manage slack during untangling or reveal additional information about the cable state $s \in \mathcal{S}$. As $s$ is difficult to estimate from the provided observation $o \in \mathcal{O}$, \algabbr{} uses a policy $\pi: \mathcal{O} \rightarrow \mathcal{A}$ built as described in Section \ref{subsec:algorithm_flow} from the components in Sections \ref{sec:IP_systems} and \ref{sec:manip-primitives} to directly predict actions to execute. Unlike prior work, \algabbr{} contains perception components that lend themselves to probabilistic interpretation and manipulation primitives that are sensitive to perception uncertainty. We note that while the distribution of states the robot encounters may have high variance, \algabbr{} is sensitive only to variance that may affect the next untangling action.
Practically, this means that for example, even if much of the cable is bunched up and occluded, as long as the path from an endpoint to the first knot is clearly visible, the algorithm can still take an untangling action with high confidence. \algabbr{} is designed to be sensitive only to uncertainty that affects the untangling process.



\subsection{Uncertainty-Aware Perception Systems} \label{sec:IP_systems}
\subsubsection{Endpoint Detection}
We train a Faster R-CNN model with a Resnet-50 FPN (Feature Pyramid Network) backbone \cite{ren2017frcnn} on 305 labeled examples to detect cable endpoints.
We discard all bounding boxes with lower than 99\% confidence, achieving an average precision and recall for endpoint detection are 86.7\% and 100\% respectively.
\subsubsection{Knot Detection}
\label{knot_det}
To identify all knots in the observable workspace, we use the same architecture as the endpoint model trained on 688 labeled images. The dataset is augmented with flip, contrast, brightness, rotation, saturation, and scale augmentations. We use a 99\% detection threshold, which achieves an average precision of 91.3\% and an average recall of 95.5\%. We analytically filter out misclassified knots by checking if a simple loop fills the bounding box. Because the model's output is dependent on the orientation of the cable, in certain cases, we use multiple observations of the underlying cable state $s$ as described in Section \ref{subsec:incr-reid-move}. This allows \algabbr{} to be sensitive to what we define as \textit{observational uncertainty}.

\subsubsection{Cable Tracing}
\begin{figure}[!ht]
    \centering
    \includegraphics[width=\linewidth]{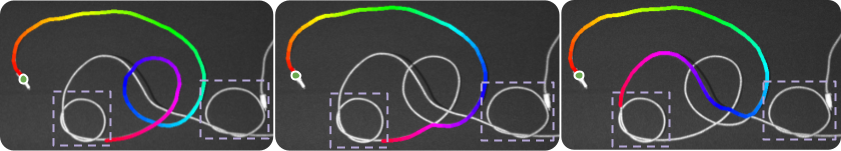}
    \vspace*{-0.25in}
    \caption{\textbf{Cable Tracing:} multiple candidate trace paths returned by the tracer to find the closest knot from the top-left endpoint. The tracer finds the correct knot in all cases, but is unclear as to which side of the knot is attached to the free endpoint, and therefore returns \texttt{TRACE\_UNCERTAIN}.}
    \label{fig:certaintrace-example}
\end{figure}

The objective of cable tracing is to follow the path of the cable from an overhead image. Given an RGB image and a start pixel (the center of an endpoint), the tracer we introduce in this work outputs a set of possible splines. It maintains a set of valid splines and iteratively expands it by exploring candidate successor points, preferring those that do not deviate sharply from the current spline's trajectory. An example of candidate trace paths is shown in Figure \ref{fig:certaintrace-example}. 

\algname{} uses this in 2 scenarios: (1) finding a knot to untangle and (2) achieving robust grasps near cable endpoints.
When used for finding a path from endpoint to knot, the tracer terminates once the trace intersects the knot within a bounding box. For grasping endpoints, the trace terminates after traveling a fixed distance along the cable. At the termination of tracing, we fit a bounding box $B_t$ to the ending points of all the final traces; if either dimension of $B_t$ is greater than 24 pixels for knot tracing and 12 pixels for endpoint tracing (which requires more precision), the traces diverge and thus $\texttt{TRACE\_UNCERTAIN}$ is returned; otherwise, $\texttt{TRACE\_CERTAIN}$ is returned. Note that while traces with very different topologies may be returned, as long as they end near the same point (Figure~\ref{fig:certaintrace-example}, left and center), the untangling-relevant uncertainty remains low.

\subsubsection{Cage-Pinch Dilation Point Selection}
We use an FCN \cite{long2015fcn} with a ResNet-34 backbone trained on 568 knot crops to output two heatmaps: one for the cage point and one for the pinch point. We augment this data with random rotations, flips, shear, and synthetically added distractor cables to improve robustness. Approximately 250 of these images are gathered during rollouts to mitigate distribution shift in a style similar to DAgger~\cite{ross2011dagger}. The input is a 3-channel image, where one of the channels contains a Gaussian heatmap around the segment of cable entering the image determined by the cable tracing algorithm. This additional input conditions the network and breaks the symmetry between the cage and pinch points. 
We train the network to predict the cage point as the first graspable point beyond the undercrossing forming the knot, and the pinch point as the place to secure the cable to create an opening for the free end to slide through. Example cage and pinch points and the perception pipeline to obtain the points are shown in Figure \ref{fig:cagepinch-example}.

The goal of this network is to model the probability $P_p(U_s)$ per grasp pixel $p$, where $U_s$ corresponds to untangling success on the cropped knot. We train an ensemble of 3 models on the same data but with different initializations. For the cage point, to sample from each of the ensemble heatmaps $h^{\mathrm{cage}} \in H^{\mathrm{cage}}$, we create a new heatmap as such: $h'^{\mathrm{cage}}_{i,j} = \min_{h^{\mathrm{cage}} \in H^{\mathrm{cage}}} h^{\mathrm{cage}}_{i, j}$. The same procedure is used for the pinch point. We return the cage and pinch point as the $\argmax_{i,j} h'^{\mathrm{cage}}_{i, j}$ and $\argmax_{i,j} h'^{\mathrm{pinch}}_{i, j}$, respectively. If $\max_{i,j} h'^{\textrm{cage}}_{i, j} \max_{r,s} h'^{\mathrm{pinch}}_{r, s}  < \kappa$ where $\kappa=0.35$ (calibrated empirically), we output \texttt{NETWORK\_UNCERTAIN}. Otherwise, we output \texttt{NETWORK\_CERTAIN}.

The above methods help reveal whether the worst-case probability (across our predictive distribution modeled by an ensemble) of untangling success is high enough to proceed with untangling the cable at the specified points. If not, the network is too uncertain in its predicted points to proceed confidently as the next action may instead tighten the knot or lead the cable into an irrecoverable state.

\subsection{Manipulation Primitives for Interactive Perception}
\label{sec:manip-primitives}
\subsubsection{Cage-Pinch Dilation}
To untangle an individual knot, the robot leverages the flexibility of the cage-pinch grippers introduced in \cite{viswanath2022autonomously} and depicted in Figure \ref{fig:splash}(2) to cage one point and pinch another point inside the knot and pull apart the arms to a distance determined by the length of the trace from the endpoint to the knot, while moving its wrist joint in a high-frequency sinusoidal motion. A major benefit of cage-pinch actions compared to cage-cage actions from \cite{viswanath2022autonomously} is the ability to better manage slack, preventing accidentally tightening another knot. Following this action, the robot lays the cable down as far forward as kinematically feasible to isolate the newly untangled portion from the remaining cable.

\begin{figure*}[!ht]
    \centering
    \includegraphics[width=0.95\textwidth]{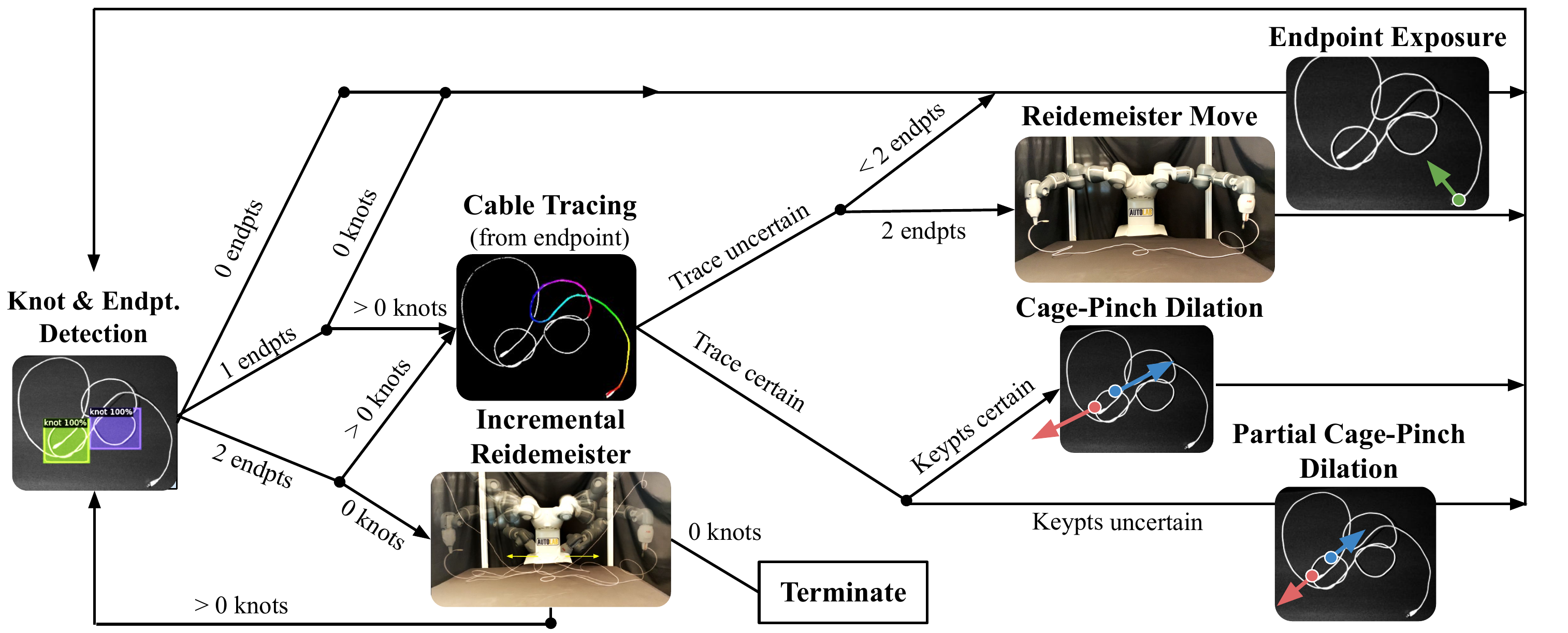}
    \vspace*{-0.1in}
    \caption{ \textbf{\algabbr{} Algorithm:}
    \algabbr{} first detects the number of knots and endpoints in the scene. If the endpoints are not visible, there is no way to verify any knot's relative position to the endpoint. This is necessary because \algabbr{} only untangles knots adjacent to an endpoint to avoid knots colliding into each other and creating irrecoverable configurations. If fewer than two endpoints and no knots are visible, the algorithm is also unable to perform a termination check as that requires performing an incremental Reidemeister, which grasps the cable at the endpoints. In both these cases, \algabbr{} performs an endpoint exposure. If two endpoints are visible and no knots are visible, \algabbr{} proceeds to the incremental Reidemeister move. If one or two endpoints are visible and there are knots in the scene, it attempts to untangle, beginning by tracing from the visible endpoint(s). Here, if it is not able to confidently trace from either endpoint to a knot, \algabbr{} performs a Reidemeister move or endpoint exposure (based on the number of endpoints visible) to increase likelihood of unambiguous traces in future steps. Otherwise, it assesses the cage-pinch network uncertainty on the predicted points. If it is confident, it proceeds with a full cage-pinch dilation. Else, it performs a partial cage-pinch dilation to disambiguate the state.}
    \label{fig:alg}
    \vspace*{-0.20in}
\end{figure*}

\subsubsection{Partial Cage-Pinch Dilation}
This primitive is similar to the Cage-Pinch Dilation, but the distance the arms move apart is fixed to 5 cm beyond their starting separation. This is meant to perturb the state and to later retry perception rather than to completely untangle a knot.


\subsubsection{Reidemeister Move}
In this primitive, the robot uses tracing to find robust grasp points slightly down the cable from the endpoints. Next, the robot moves both arms outward horizontally and up, lifting the cable off the workspace, allowing for loops to fall away. Compared to prior work, we add (1) the vertical component of the action, forming a large letter ``U" with the cable, and (2) the wide lay-out action, which places the cable on the workspace in a ``U" shape.

\subsubsection{Incremental Reidemeister Move}
\label{subsec:incr-reid-move}
This primitive performs the exact same motions as a Reidemeister move, but uses a multi-stage, perception-based approach where the cable is observed at certain waypoints along the action. We use our knot detection network at these intermediate points to determine whether any knots remain. This can be interpreted as ensembling via perturbation of the observation of the same underlying cable topology. Being sensitive to observational uncertainty allows us to eliminate the time-consuming physical tracing action used in \cite{viswanath2022autonomously} for termination.

\subsubsection{Exposure Action}
When one or more endpoints are missing for an action, we uniformly at random sample a segment of the cable leaving the reachable workspace and pull it towards the center of the workspace. We do this also for unreachable knots that we wish to act on.

\subsection{\algname{} Algorithm}
\label{subsec:algorithm_flow}
\algname{} ties together the aforementioned perception components and manipulation primitives to untangle cables. \algabbr{} alternates between the perception and manipulation components, using uncertainty from the former to determine whether  to untangle or disambiguate the cable state. The algorithm is covered in detail in Figure \ref{fig:alg}.


\section{Experiments}
\label{sec:expts}
\label{subsec:workspace_definition}
\subsection{Experimental Setup}

For our experiments, we use the bimanual ABB YuMi robot with an overhead Phoxi camera, operating on a black foam-padded workspace of width $1.0\,m$ and depth $0.75\,m$. Due to hardware constraints, we slightly extend the workspace with cardboard (by 0.1 meters on either side) not previously present in SGTM 1.0, but this does not make a significant difference as the cable mostly remains in the original foam-padded workspace. We use a $2.7\,m$-long white, braided electrical cable with USB adapters on both ends.

\begin{figure}[!ht]
    \vspace*{-0.18in}
    \centering
    \includegraphics[width=1.0\linewidth]{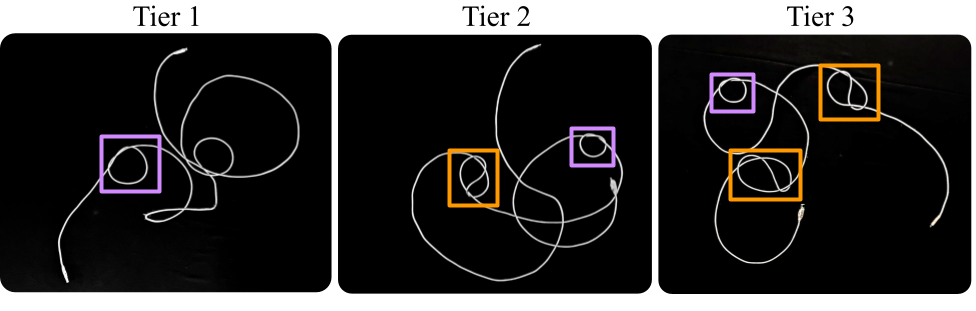}
    \vspace*{-0.19in}
    \caption{\textbf{Example starting configurations for all 3 tiers:} Overhand knots are outlined in purple and figure-8 knots are outlined in orange.}
    \label{fig:start-configs}
    \vspace*{-0.18in}
\end{figure}

We evaluate \algabbr{} on 3 tiers of difficulty:
\begin{enumerate}
    \item \textbf{Tier 1:} A cable with 1 overhand or figure-8 knot.
    \item \textbf{Tier 2:} A cable with 2 overhand and/or figure-8 knots. 
    \item \textbf{Tier 3:} A cable with 3 overhand and/or figure-8 knots. 
\end{enumerate}
The knots in all tiers are evaluated equally in both loose and dense configurations and evenly across positions along the cable (closer to an endpoint vs. closer to the middle). Example start configurations are shown in Figure \ref{fig:start-configs}. The cable is initialized by laying the knot(s) flat on the workspace, raising the endpoints as high as possible without lifting the knot(s), and then dropping the endpoints.
We enforce a time limit of 15 minutes for all tiers. Note that the cable initialization procedures in Tiers 1 and 2 of this work are \textit{exactly} the same as Tiers 1 and 2 in SGTM 1.0, the prior state-of-the-art \cite{viswanath2022autonomously}.

For each tier, we report the average time to fully untangle the cable (for the rollouts that succeed in doing so) as well as the average time to correctly report that the cable is untangled (for the rollouts that succeed in doing so). Additionally, we report the success rates for untangling alone and untangling with termination detection. For Tiers 2 and 3, we present ablation results where \algabbr{}(-U) represents \algabbr{} with the uncertainty-based components removed. For Tiers 1 and 2, we also report the speedup of \algabbr{} and \algabbr{}(-U) from SGTM 1.0.

\begin{figure}[ht!]
\centering
\includegraphics[width=1.0\linewidth]{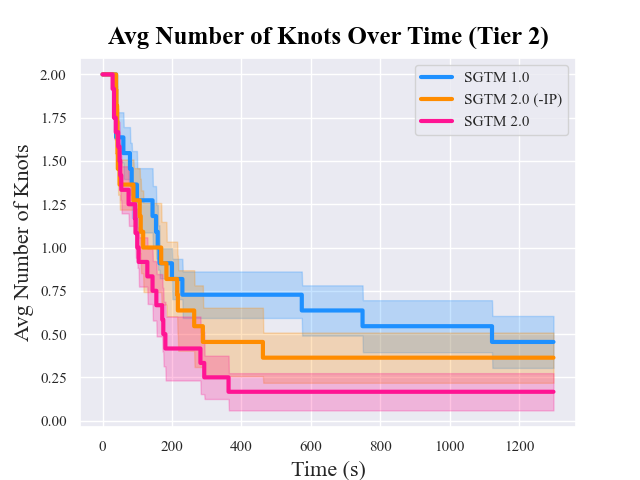}
\vspace*{-0.25in}
\caption{\textbf{Average Knots over Time (Tier 2)}: \algabbr{} most quickly reduces the number of knots on average. \algabbr{}(-U) is less successful in untangling over the given time period than \algabbr{}. Further, performance of even \algabbr{}(-U) exceeds that of the prior state-of-the-art, SGTM 1.0, at 15 minutes, showing the effectiveness of improved untangling primitives such as cage-pinch dilations introduced in \algabbr{}.}
\vspace*{-0.20in}
\label{fig:Tier2_knots_over_time}
\end{figure}

\begin{table*}[!t]
    \centering
    \newcolumntype{?}{!{\vrule width 1.75pt}}
    \caption{Results from Physical Experiments (84 total trials)}
    \setlength\tabcolsep{4pt}
    \begin{tabular}{|c?c|c?c|c|c?c|c|} \hline 
    \multicolumn{1}{|c?}{} &
      \multicolumn{2}{|c?}{\cellcolor{MistyRose}{{\textbf{Tier 1}}}} & \multicolumn{3}{|c?}{\cellcolor{AliceBlue}{\textbf{Tier 2}}} & \multicolumn{2}{|c|}{\cellcolor{Honeydew}{\textbf{Tier 3}}}\\ \cline{2-8} 
    \multicolumn{1}{|c?}{} & {SGTM 1.0} & {\algabbr{}} & {SGTM 1.0} & {\algabbr{}(-U)} & \algabbr{} & {\algabbr{}(-U)} & {\algabbr{}} \\ \hline
    Knot 1 Success Rate & {8/12} & {\textbf{10/12}} & {10/12} & {10/12} & {\textbf{12/12}} & {12/12} & {12/12} \\ \hline
    Knot 2 Success Rate & {-} & {-} & 6/12 & 7/12 & \textbf{10/12} & 9/12 & 9/12 \\ \cline{1-8} 
    Knot 3 Success Rate & {-} & {-} & - & - & - & 3/12 & 3/12 \\
    \cline{1-8} 
    Verification Rate & {3/12} & {\textbf{7/12}} & 1/12 & 5/12 & \textbf{7/12} & 0/12 & \textbf{2/12}\\
    \cline{1-8}
    Avg. \# of Actions & {\textbf{5.0$\pm$1.5}} & {5.6$\pm$1.7} & 12.0 & \textbf{6.0$\pm$1.2} & 7.7$\pm$1.6 & N/A & \textbf{12.0$\pm$0.0} \\
    \cline{1-8}
    Avg. Knot 1 Time (s) & {189.8$\pm$69.4} & {\textbf{53.1$\pm$7.5 (3.6x)}} & 93.3$\pm$16.2 & 128.6$\pm$41.9 (0.7x) & \textbf{75.6$\pm$21.0 (1.2x)} & 88.7$\pm$25.5 & \textbf{69.8$\pm$15.1} \\
    \cline{1-8}
    Avg. Knot 2 Time (s) & {-} & {-} & 586.4$\pm$165.3 & \textbf{160.9$\pm$26.4 (3.6x)} & 180.9$\pm$26.2 (3.2x) & \textbf{177.3$\pm$28.3} & 233.1$\pm$57.6\\
    \cline{1-8}
    Avg. Knot 3 Time (s) & {-} & {-} & - & - & - & \textbf{417.7$\pm$104.1} & 476.7$\pm$160.1\\
    \cline{1-8}
    Avg. Verif. Time (s) & {330.8$\pm$127.8} & {\textbf{295.9$\pm$61.4 (1.1x)}} & 1079.0 & \textbf{359.6$\pm$74.6 (3.0x)} & 406.9$\pm$22.0 (2.7x) & N/A & \textbf{704.5$\pm$13.5}\\
    \cline{1-8}
    \multicolumn{1}{|c?}{Failures} & {-} & {(A) 2, (B) 1,} & - & (A) 1, (B) 0,& (A) 2, (B) 1,& (A) 1, (B) 4,& (A) 5, (B) 2,\\
    & {} & {(C) 1, (D) 1} & & (C) 5, (D) 1 & (C) 1, (D) 1 & (C) 6, (D) 1 & (C) 2, (D) 1\\
    \cline{1-8}
    \hline 
    \end{tabular}
    \label{tab:results_all}
    \vspace{-0.2in}
\end{table*}

\subsection{Results}
Results show that across Tiers 1 and 2, \algabbr{} outperforms SGTM 1.0 not only on untangling and verification success rate, but also achieves statistically significant speedups in the untangling time in Tier 1, untangling time for both knots in Tier 2, and verification time in Tier 2. Tier 3 was unachievable in SGTM 1.0 due to algorithmic constraints, but is now possible with SGTM 2.0, which achieves 9/12 successes in untangling 2 out of the 3 knots in Tier 3. In \ref{subsec:failure_modes}, we discuss difficulties that result in failures, leading to 3/12 untangling success on all 3 knots in Tier 3.

\subsection{Failure Modes}\label{subsec:failure_modes}
Across all 3 tiers, we observe the following failure modes.

\textbf{(A) Timeout, unable to determine termination:} This is the most common failure case of SGTM 2.0 across all tiers, especially in Tier 3. Causes include:
\begin{enumerate}
    \item The system inadvertently manipulates the cable into a state that is difficult to perceive and manipulate, most common in Tier 3 due to higher complexity and a higher chance that a rare failure in disambiguation may accidentally tighten or complicate a knot.
    \item A substantial portion of the cable leaves the workspace. The system repeatedly attempts exposure actions, but due to the weight of the endpoint or large mass of cable, the cable continually slips back down into its prior configuration.
    \item Though the entire cable may enter a difficult configuration, the algorithm slowly disambiguates it and given more time, may have untangled and terminated.
\end{enumerate}

\textbf{(B) Cable or knot leaves observable/reachable workspace:} This is the next most common failure mode of \algabbr{}. While performing a cage-pinch dilation or Reidmeister move, one gripper may miss the grasp, causing the entire cable to slide to one side of the workspace and fall off entirely and irrecoverably. This failure mode shows that slack management can be improved in future work.

\textbf{(C) False termination due to missed knot detection:} False termination is the most common failure mode in \algabbr{}(-U), mostly resolved by \algabbr{} with uncertainty-based components. This failure also occurs in \algabbr{}, largely due to rarer cases where the knotted portion inadvertently lands outside the observable workspace during an incremental Reidemeister move, causing early termination. This can be addressed with improved motion primitives.

\textbf{(D) Irrecoverable YuMi system error:} These relatively rare issues result from the YuMi losing connection to the computer running the algorithm and freezing.

\subsection{Ablations}
We run ablations on Tier 2 and Tier 3 to compare the performance of \algabbr{} to the performance of \algabbr{}(-U), which uses the exact same algorithm as \algabbr{}, but with the following uncertainty-based components removed:
\begin{itemize}
    \item Reidemeister move due to tracing uncertainty.
    \item Ensemble network for keypoint predictions and partial cage-pinch dilation in the case of ensemble uncertainty in the cage-pinch dilation network.
    \item Intermediate views for incremental Reidemeister move.
\end{itemize}

We find, as shown in table \ref{tab:results_all}, that \algabbr{}(-U) achieves a lower success rate than \algabbr{} on Tier 2. While \algabbr{} and \algabbr{}(-U) achieve the same success rate on Tier 3, the main failure case for \algabbr{} is timeout as higher complexity cases tend to require more time to disambiguate and untangle. In comparison, the main failure case for \algabbr{}(-U) are false termination. In fact, the most common failure case in \algabbr{}(-U) across Tier 2 and 3 is false termination, suggesting that sensitivity to observational uncertainty may be important for higher performance. Another implicit failure that results in more false terminations is over-tightening of knots to a diameter $\leq 3 cm$. If the ensemble cage-pinch network has low confidence, \algabbr{} performs a partial cage-pinch dilation rather than a full cage-pinch dilation, preventing over-tightening knots in the case of poorly predicted cage-pinch points. Additionally, if the trace to a knot is uncertain, the Reidemeister move disambiguates the cable state, preventing poor grasps that may tighten or complicate the knots. The ablations suggest that these uncertainty-based primitives may prevent the over-tightening of knots and thus reduce false terminations.

\section{Conclusion}

In this paper, we significantly extend our prior work on SGTM 1.0 to present \algabbr{}, with novel uncertainty-based and active perception actions. \algabbr{} achieves an average untangling success rate of 83\% and average rollout time of 351 seconds on cables with 1 or 2 knots, outperforming the prior state-of-the-art, SGTM 1.0, in untangling success by 43\% and in untangling speed by 200\%. We find that introducing interactive perception -- actively manipulating the cable to facilitate perception -- improves untangling success on complex cases by 21\%.

In future work, we will explore adding interactive perception and uncertainty modeling into other parts of the pipeline as well as generalization to more knot and cable types provided sufficient data. Another potential direction involves eliminating depth sensing and using interactive perception to achieve robust grasps with just RGB data.








\bibliographystyle{IEEEtran}
\bibliography{IEEEabrv,references}

\begin{thebibliography}{10}
\providecommand{\url}[1]{#1}
\csname url@samestyle\endcsname
\providecommand{\newblock}{\relax}
\providecommand{\bibinfo}[2]{#2}
\providecommand{\BIBentrySTDinterwordspacing}{\spaceskip=0pt\relax}
\providecommand{\BIBentryALTinterwordstretchfactor}{4}
\providecommand{\BIBentryALTinterwordspacing}{\spaceskip=\fontdimen2\font plus
\BIBentryALTinterwordstretchfactor\fontdimen3\font minus
  \fontdimen4\font\relax}
\providecommand{\BIBforeignlanguage}[2]{{%
\expandafter\ifx\csname l@#1\endcsname\relax
\typeout{** WARNING: IEEEtran.bst: No hyphenation pattern has been}%
\typeout{** loaded for the language `#1'. Using the pattern for}%
\typeout{** the default language instead.}%
\else
\language=\csname l@#1\endcsname
\fi
#2}}
\providecommand{\BIBdecl}{\relax}
\BIBdecl

\bibitem{mayer2008system}
H.~Mayer, F.~Gomez, D.~Wierstra, I.~Nagy, A.~Knoll, and J.~Schmidhuber, ``A
  system for robotic heart surgery that learns to tie knots using recurrent
  neural networks,'' \emph{Advanced Robotics}, vol.~22, no. 13-14, pp.
  1521--1537, 2008.

\bibitem{sanchez2018robotic}
J.~Sanchez, J.-A. Corrales, B.-C. Bouzgarrou, and Y.~Mezouar, ``Robotic
  manipulation and sensing of deformable objects in domestic and industrial
  applications: a survey,'' \emph{The International Journal of Robotics
  Research}, vol.~37, no.~7, pp. 688--716, 2018.

\bibitem{van2010superhuman}
J.~Van Den~Berg, S.~Miller, D.~Duckworth, H.~Hu, A.~Wan, X.-Y. Fu, K.~Goldberg,
  and P.~Abbeel, ``Superhuman performance of surgical tasks by robots using
  iterative learning from human-guided demonstrations,'' in \emph{2010 IEEE
  International Conference on Robotics and Automation}.\hskip 1em plus 0.5em
  minus 0.4em\relax IEEE, 2010, pp. 2074--2081.

\bibitem{viswanath2022autonomously}
V.~Viswanath, K.~Shivakumar, J.~Kerr, B.~Thananjeyan, E.~Novoseller,
  J.~Ichnowski, A.~Escontrela, M.~Laskey, J.~E. Gonzalez, and K.~Goldberg,
  ``Autonomously untangling long cables,'' \emph{Robotics: Science and Systems
  (RSS)}, 2022.

\bibitem{bohg2017interactive}
J.~Bohg, K.~Hausman, B.~Sankaran, O.~Brock, D.~Kragic, S.~Schaal, and G.~S.
  Sukhatme, ``Interactive perception: Leveraging action in perception and
  perception in action,'' \emph{IEEE Transactions on Robotics}, vol.~33, no.~6,
  pp. 1273--1291, 2017.

\bibitem{seita2019deep}
D.~Seita, A.~Ganapathi, R.~Hoque, M.~Hwang, E.~Cen, A.~K. Tanwani,
  A.~Balakrishna, B.~Thananjeyan, J.~Ichnowski, N.~Jamali \emph{et~al.}, ``Deep
  imitation learning of sequential fabric smoothing from an algorithmic
  supervisor,'' 2020.

\bibitem{seita2020learning}
D.~Seita, P.~Florence, J.~Tompson, E.~Coumans, V.~Sindhwani, K.~Goldberg, and
  A.~Zeng, ``Learning to rearrange deformable cables, fabrics, and bags with
  goal-conditioned transporter networks,'' 2021.

\bibitem{nair2017combining}
A.~Nair, D.~Chen, P.~Agrawal, P.~Isola, P.~Abbeel, J.~Malik, and S.~Levine,
  ``Combining self-supervised learning and imitation for vision-based rope
  manipulation,'' in \emph{2017 IEEE Int. Conf. on Robotics and Automation
  (ICRA)}.\hskip 1em plus 0.5em minus 0.4em\relax IEEE, 2017, pp. 2146--2153.

\bibitem{matas2018sim}
J.~Matas, S.~James, and A.~J. Davison, ``Sim-to-real reinforcement learning for
  deformable object manipulation,'' in \emph{Conference on Robot
  Learning}.\hskip 1em plus 0.5em minus 0.4em\relax PMLR, 2018, pp. 734--743.

\bibitem{wu2019learning}
Y.~Wu, W.~Yan, T.~Kurutach, L.~Pinto, and P.~Abbeel, ``Learning to manipulate
  deformable objects without demonstrations,'' \emph{Robotics: Science and
  Systems (RSS)}, 2019.

\bibitem{lee2020learning}
R.~Lee, D.~Ward, A.~Cosgun, V.~Dasagi, P.~Corke, and J.~Leitner, ``Learning
  arbitrary-goal fabric folding with one hour of real robot experience,''
  \emph{Conference on Robot Learning}, 2020.

\bibitem{seita2020deep}
D.~Seita, A.~Ganapathi, R.~Hoque, M.~Hwang, E.~Cen, A.~K. Tanwani,
  A.~Balakrishna, B.~Thananjeyan, J.~Ichnowski, N.~Jamali \emph{et~al.}, ``Deep
  imitation learning of sequential fabric smoothing from an algorithmic
  supervisor,'' in \emph{2020 IEEE/RSJ International Conference on Intelligent
  Robots and Systems (IROS)}.\hskip 1em plus 0.5em minus 0.4em\relax IEEE,
  2020, pp. 9651--9658.

\bibitem{seita2021learning}
D.~Seita, P.~Florence, J.~Tompson, E.~Coumans, V.~Sindhwani, K.~Goldberg, and
  A.~Zeng, ``Learning to rearrange deformable cables, fabrics, and bags with
  goal-conditioned transporter networks,'' in \emph{2021 IEEE International
  Conference on Robotics and Automation (ICRA)}.\hskip 1em plus 0.5em minus
  0.4em\relax IEEE, 2021, pp. 4568--4575.

\bibitem{yan2020learning}
\BIBentryALTinterwordspacing
W.~Yan, A.~Vangipuram, P.~Abbeel, and L.~Pinto, ``Learning predictive
  representations for deformable objects using contrastive estimation,'' in
  \emph{Proceedings of the 2020 Conference on Robot Learning}, ser. Proceedings
  of Machine Learning Research, J.~Kober, F.~Ramos, and C.~Tomlin, Eds., vol.
  155.\hskip 1em plus 0.5em minus 0.4em\relax PMLR, 16--18 Nov 2021, pp.
  564--574. [Online]. Available:
  \url{https://proceedings.mlr.press/v155/yan21a.html}
\BIBentrySTDinterwordspacing

\bibitem{sundaresan2020learning}
P.~Sundaresan, J.~Grannen, B.~Thananjeyan, A.~Balakrishna, M.~Laskey, K.~Stone,
  J.~E. Gonzalez, and K.~Goldberg, ``Learning rope manipulation policies using
  dense object descriptors trained on synthetic depth data,'' in \emph{2020
  IEEE International Conference on Robotics and Automation (ICRA)}.\hskip 1em
  plus 0.5em minus 0.4em\relax IEEE, 2020, pp. 9411--9418.

\bibitem{ganapathi2021learning}
A.~Ganapathi, P.~Sundaresan, B.~Thananjeyan, A.~Balakrishna, D.~Seita,
  J.~Grannen, M.~Hwang, R.~Hoque, J.~E. Gonzalez, N.~Jamali \emph{et~al.},
  ``Learning dense visual correspondences in simulation to smooth and fold real
  fabrics,'' in \emph{2021 IEEE International Conference on Robotics and
  Automation (ICRA)}.\hskip 1em plus 0.5em minus 0.4em\relax IEEE, 2021, pp.
  11\,515--11\,522.

\bibitem{hoque2020visuospatial}
R.~Hoque, D.~Seita, A.~Balakrishna, A.~Ganapathi, A.~K. Tanwani, N.~Jamali,
  K.~Yamane, S.~Iba, and K.~Goldberg, ``Visuospatial foresight for multi-step,
  multi-task fabric manipulation,'' \emph{Robotics: Science and Systems (RSS)},
  2020.

\bibitem{zhang2021robots}
H.~Zhang, J.~Ichnowski, D.~Seita, J.~Wang, H.~Huang, and K.~Goldberg, ``Robots
  of the lost arc: Self-supervised learning to dynamically manipulate
  fixed-endpoint cables,'' in \emph{2021 IEEE International Conference on
  Robotics and Automation (ICRA)}.\hskip 1em plus 0.5em minus 0.4em\relax IEEE,
  2021, pp. 4560--4567.

\bibitem{limr2s2r2022}
V.~Lim, H.~Huang, L.~Y. Chen, J.~Wang, J.~Ichnowski, D.~Seita, M.~Laskey, and
  K.~Goldberg, ``Real2sim2real: Self-supervised learning of physical
  single-step dynamic actions for planar robot casting,'' in \emph{2022
  International Conference on Robotics and Automation (ICRA)}, 2022, pp.
  8282--8289.

\bibitem{avigal2022speedfolding}
Y.~Avigal, L.~Berscheid, T.~Asfour, T.~Kr{\"o}ger, and K.~Goldberg,
  ``Speedfolding: Learning efficient bimanual folding of garments,''
  \emph{International Conference on Intelligent Robots and Systems (IROS)
  2022}, 2022.

\bibitem{chenfling2022}
L.~Y. Chen, H.~Huang, E.~Novoseller, D.~Seita, J.~Ichnowski, M.~Laskey,
  R.~Cheng, T.~Kollar, and K.~Goldberg, ``Efficiently learning single-arm fling
  motions to smooth garments,'' in \emph{International Symposium on Robotics
  Research}, 2022.

\bibitem{florence2018dense}
P.~R. Florence, L.~Manuelli, and R.~Tedrake, ``Dense object nets: Learning
  dense visual object descriptors by and for robotic manipulation,'' 2018.

\bibitem{ganapathi2020learning}
A.~Ganapathi, P.~Sundaresan, B.~Thananjeyan, A.~Balakrishna, D.~Seita,
  J.~Grannen, M.~Hwang, R.~Hoque, J.~E. Gonzalez, N.~Jamali \emph{et~al.},
  ``Learning to smooth and fold real fabric using dense object descriptors
  trained on synthetic color images,'' 2021.

\bibitem{wang2019learning2}
A.~Wang, T.~Kurutach, K.~Liu, P.~Abbeel, and A.~Tamar, ``Learning robotic
  manipulation through visual planning and acting,'' \emph{Robotics: Science
  and Systems (RSS)}, 2019.

\bibitem{lin2022learning}
X.~Lin, Y.~Wang, Z.~Huang, and D.~Held, ``Learning visible connectivity
  dynamics for cloth smoothing,'' in \emph{Conference on Robot Learning}.\hskip
  1em plus 0.5em minus 0.4em\relax PMLR, 2022, pp. 256--266.

\bibitem{viswanath2021disentangling}
V.~Viswanath, J.~Grannen, P.~Sundaresan, B.~Thananjeyan, A.~Balakrishna,
  E.~Novoseller, J.~Ichnowski, M.~Laskey, J.~E. Gonzalez, and K.~Goldberg,
  ``Disentangling dense multi-cable knots,'' 2021.

\bibitem{grannen2020untangling}
J.~Grannen, P.~Sundaresan, B.~Thananjeyan, J.~Ichnowski, A.~Balakrishna,
  M.~Hwang, V.~Viswanath, M.~Laskey, J.~E. Gonzalez, and K.~Goldberg,
  ``Untangling dense knots by learning task-relevant keypoints,''
  \emph{Conference on Robot Learning}, 2020.

\bibitem{lui2013tangled}
W.~H. Lui and A.~Saxena, ``Tangled: Learning to untangle ropes with {RGB-D}
  perception,'' in \emph{2013 IEEE/RSJ Int. Conf. on Intelligent Robots and
  Systems}.\hskip 1em plus 0.5em minus 0.4em\relax IEEE, 2013, pp. 837--844.

\bibitem{sundaresan2021untangling}
P.~Sundaresan, J.~Grannen, B.~Thananjeyan, A.~Balakrishna, J.~Ichnowski,
  E.~Novoseller, M.~Hwang, M.~Laskey, J.~E. Gonzalez, and K.~Goldberg,
  ``Untangling dense non-planar knots by learning manipulation features and
  recovery policies,'' 2021.

\bibitem{bajcsy1988active}
R.~Bajcsy, ``Active perception,'' \emph{Proceedings of the IEEE}, vol.~76,
  no.~8, pp. 966--1005, 1988.

\bibitem{bajcsy2018revisiting}
R.~Bajcsy, Y.~Aloimonos, and J.~K. Tsotsos, ``Revisiting active perception,''
  \emph{Autonomous Robots}, vol.~42, no.~2, pp. 177--196, 2018.

\bibitem{goldberg1984active}
K.~Y. Goldberg and R.~Bajcsy, ``Active touch and robot perception,''
  \emph{Cognition and Brain Theory}, vol.~7, no.~2, pp. 199--214, 1984.

\bibitem{novkovic2020object}
T.~Novkovic, R.~Pautrat, F.~Furrer, M.~Breyer, R.~Siegwart, and J.~Nieto,
  ``Object finding in cluttered scenes using interactive perception,'' in
  \emph{2020 IEEE International Conference on Robotics and Automation
  (ICRA)}.\hskip 1em plus 0.5em minus 0.4em\relax IEEE, 2020, pp. 8338--8344.

\bibitem{tsikos1988segmentation}
C.~J. Tsikos and R.~K. Bajcsy, ``Segmentation via manipulation,''
  \emph{Technical Reports (CIS)}, p. 694, 1988.

\bibitem{danielczuk2019mechanical}
M.~Danielczuk, A.~Kurenkov, A.~Balakrishna, M.~Matl, D.~Wang,
  R.~Mart{\'\i}n-Mart{\'\i}n, A.~Garg, S.~Savarese, and K.~Goldberg,
  ``Mechanical search: Multi-step retrieval of a target object occluded by
  clutter,'' in \emph{2019 International Conference on Robotics and Automation
  (ICRA)}.\hskip 1em plus 0.5em minus 0.4em\relax IEEE, 2019, pp. 1614--1621.

\bibitem{willimon2011classification}
B.~Willimon, S.~Birchfield, and I.~Walker, ``Classification of clothing using
  interactive perception,'' in \emph{2011 IEEE International Conference on
  Robotics and Automation}.\hskip 1em plus 0.5em minus 0.4em\relax IEEE, 2011,
  pp. 1862--1868.

\bibitem{ren2017frcnn}
S.~Ren, K.~He, R.~Girshick, and J.~Sun, ``Faster r-cnn: Towards real-time
  object detection with region proposal networks,'' \emph{IEEE Transactions on
  Pattern Analysis and Machine Intelligence}, vol.~39, no.~6, pp. 1137--1149,
  2017.

\bibitem{long2015fcn}
J.~Long, E.~Shelhamer, and T.~Darrell, ``Fully convolutional networks for
  semantic segmentation,'' \emph{CVPR}, 2015.

\bibitem{ross2011dagger}
S.~Ross, G.~J. Gordon, and J.~A. Bagnell, ``A reduction of imitation learning
  and structured prediction to no-regret online learning,'' 2011.

\end{thebibliography}



\end{document}